# PEACH: a sentence-aligned Parallel English–Arabic Corpus for Healthcare

Rania Al-Sabbagh[1]

**Abstract**
This paper introduces peach, a sentence-aligned parallel English–Arabic corpus of healthcare texts encompassing patient information leaflets and educational materials. The corpus contains 51,671 parallel sentences, totalling approximately 590,517 English and 567,707 Arabic word tokens. Sentence lengths vary between 9.52 and 11.83 words on average. As a manually aligned corpus, peach is a gold-standard corpus, aiding researchers in contrastive linguistics, translation studies, and natural language processing. It can be used to derive bilingual lexicons, adapt large language models for domain-specific machine translation, evaluate user perceptions of machine translation in healthcare, assess patient information leaflets and educational materials' readability and lay-friendliness, and as an educational resource in translation studies. peach is publicly accessible.[2]

## 1. Introduction

Parallel corpora (i.e., collections of texts and their translations) are essential resources to compare and contrast languages (Hasselgård, 2020; Oliver and Mikeleni, 2020; and Le Bruyn *et al*., 2022), create bilingual lexicons (Marchisio *et al*., 2021), develop machine translation models (Han, 2018), fine-tune large language models for domain-specific translation (Moslem *et al*., 2023), assess end-user attitudes toward machine translation (Kaspereˇ *et al*., 2021), develop and evaluate automatic post-editing models (Do Carmo *et al*., 2020), and train human translators in translation classes (Liu, 2020).

---

[1] Department of Foreign Languages, University of Sharjah, United Arab Emirates, PO Box 27272, United Arab Emirates.
*Correspondence to*: Rania Al-Sabbagh, *e-mail*: rmalsabbagh@sharjah.ac.ae

[2] See: https://data.mendeley.com/datasets/5k6yrrhng7/1.

This paper presents peach, a sentence-aligned parallel corpus of healthcare text, which includes 517 English and Arabic patient information leaflets and educational materials.

The patient information leaflets featured in peach were sourced from the Saudi Drugs Information System[3] (sdis), affiliated with the Saudi Food and Drug Authority (sfda). The Saudi Council of Ministers established the authority in 2003 to monitor and regulate food products, pharmaceuticals, medical devices, laboratories, cosmetics and tobacco. Available in English and Arabic, the sfda website offers several English and Arabic patient information leaflets. Also referred to as medication package inserts, consumer medicine information, and medical brochures, these leaflets provide information on medications, detailing their ingredients, intended effects, potential side effects, recommended dosages, and guidance on what to do if side effects occur.

Translating the leaflets into Arabic is not explicitly detailed, leaving ambiguity about whether the translations were manually created or derived from machine translation followed by post-editing. According to the Gulf Cooperation Council (gcc) guidelines, which the sfda follows, 'labeling information [is] required to be translated into [the] Arabic language', a rule that is not applicable to medicines used only in hospitals (Gulf Health Council, 2021: 44). In addition, the sfda's instructions for the presentation of patient information leaflets and labelling for herbal and health products detail the structure and content of the leaflets, emphasising that '[a] leaflet has to be professionally translated, proofread, and written in a language that the general public can understand' (Saudi Food & Drug Authority, 2017: 22). The sfda further clarifies that 'submitted patient information leaflets will not be considered unless they meet the previous conditions' (Saudi Food & Drug Authority, 2017: 22). This guidance implies that the translation responsibility lies with the pharmaceutical companies, which must submit their translations for sfda approval. Whilst the precise translation process is not detailed, the resulting patient information leaflets, officially published by a respected organisation in the Arab world, are deemed valid reference translations.

The patient educational materials in peach were collected from the Health Information Translations (hit) project[4] sponsored by the Central Ohio Hospital Council, Mount Carmel, Nationwide Children, Ohio Health, and the Ohio State University in the United States. The project aimed to provide plain language health guidelines for healthcare professionals and others working in communities with limited English proficient populations. The materials are offered in multiple languages, including Arabic, and cover various healthcare topics, such as diagnostic tests, disaster preparedness, home care, safety and rehabilitation. According to Health Information Translations (n.d.), the translations were done manually by an authorised translation firm in the United States (US), a corporate American Translators Association member.

---

[3] See: https://sdi.sfda.gov.sa/.
[4] See: https://www.healthinfotranslations.org/.

| Precautions | الاحتياطات |
|---|---|
| • Keep the splint away from open flames because it will burn. | • أبعد الجبيرة عن ألسنة اللهب المكشوفة لأنها ستحترق. |
| • Keep the splint away from heat, water heaters, or prolonged sunlight, such as in a hot, closed car. Excessive heat will cause the splint to change shape. | • أبعد الجبيرة عن الحرارة أو سخانات المياه أو تجنب تعريضها لضوء الشمس لفترات طويلة كما في حالة وضعها داخل سيارة مغلقة في جو حار، حيث إن الحرارة الزائدة قد تؤدي إلى تغيير شكل الجبيرة. |
| • Put your hand and forearm up on pillows or a wedge while in bed or when sitting. | • ارفع يدك والساعد لأعلى على وسائد أو مسند أثناء النوم على السرير أو أثناء الجلوس. |

**Table 1**: An excerpt from the peach corpus showcasing a list of imperative verbs. It is derived from a patient educational material entitled 'How to wear and care for your splint'.

All the translators involved in the translation process had at least a bachelor's degree in their field and underwent tests before working on the project.

Patient information leaflets and educational materials are distinct textual genres that stand out from others in English–Arabic parallel corpora, like newspaper articles, film subtitles, and tweets. Ornia (2016) describes them as heterogeneous genres that serve a mix of expository and instructive functions with common and specialised communicative tones. They are rich in medical terminology, encompassing categories such as chemicals (for instance, cetostearyl alcohol, methylparaben and mefenamic acid), symptoms and diseases (like ulcerative colitis, peri-oral dermatitis and hyperlipidemia), and medical devices and procedures (such as barium enema and appendectomy). Since they serve an instructive function, patient information leaflets and educational materials excessively use imperative verbs, as illustrated in the excerpt from the peach corpus in Table 1. Another distinctive feature of patient information leaflets and educational materials is the excessive use of enumerated lists, where the items on the list could be phrases (see Table 2) or parts of compound and complex clauses (see Table 3).

## 2.    Related work

Algabbani *et al*. (2022), Lin *et al*. (2020), Malik *et al*. (2019) and Sharkas (2019) researched the readability and user-friendliness of patient information leaflets and health guidelines in Arabic, aiming to evaluate their accessibility to the Arabic-speaking population. Specifically, Algabbani *et al*. (2022) analysed the readability of anti-hypertensive medication leaflets from the Saudi Drug Information System, the same source as the peach corpus. Whilst the number of leaflets analysed was not specified, the study utilised sentence length to assess the readability of these Arabic translations. The findings revealed that 83 percent of the leaflets for generic medications and 68 percent for brand-name medications exceeded the recommended readability threshold. Consequently, Algabbani *et al*. suggested establishing guidelines to enhance the readability of these leaflets and recommended that manufacturers undergo a readability assessment before submitting their leaflets to the sfda.

| Common side effects (may affect up to 1 in 10 people): | الآثار الجانبية الشَائعة (قَد تُؤثِر على ما يصِل إلى شخص واحد من بين كل 10 أشخاص): |
|---|---|
| • yeast infection | • عدوى فطرية |
| • abnormal lab test (positive direct Coombs) | • نتائج غير طبيعية للاختبارات المعملية (نتائج إيجابية مباشرة لاختبار كومبس) |
| • prolonged blood clotting time (activated partial thromboplastin time prolonged) | • اطالة زمن تخثر الدم (إطالة زمن الثرومبوبلاستين الجزئي المفعّل) |
| • decrease in blood protein | • انخفاض في البروتين بالدم |
| • upset stomach | • اضطراب بالمعدة |
| • increase in blood liver enzymes | • ارتفاع بإنزيمات الكبد بالدم |
| • abnormal kidney blood tests | • نتائج غير طبيعية باختبارات الكلى بالدم |
| • fever | • حمى |
| • injection site reaction | • تفاعل بموضع الحقن |

**Table 2**: An excerpt from the peach corpus showcasing a list of noun phrases. It is derived from a patient information leaflet for a drug named Piperacillin.

| Talk to your doctor, nurse, or pharmacist before you have your medicine if: | تحدث مع طبيبك، الممرض او الصيدلي قبل استخدام دوائك إذا: |
|---|---|
| • You are 60 years of age or older | • كنت تبلغ من العمر 60 سنة أو أكبر |
| • You are using corticosteroids, sometimes called steroids (see section 'Other medicines and Levonic') | • كنت تستخدم الستيرويدات القشرية، تسمى أحياناً بالستيرويدات (انظر "قسم الأدوية الأخرى وليفونك") |
| • You have ever had a fit (seizure) | • عانيت مسبقاً من نوبات (نوبات تشنجية) |
| • You have had damage to your brain due to a stroke or other brain injury | • سبق لك أن عانيت من تلف في دماغك بسبب سكتة دماغية أو إصابة أخرى في الدماغ |
| • You have kidney problems | • كنت تعاني من مشاكل في الكلى |
| • You have a 'glucose 6 – phosphate dehydrogenase (G6PD) deficiency'. | • كنت مصاباً بشيء يعرف باسم 'عوز نازعة الهيدروجين جلوكوز 6 – فوسفات.' |

**Table 3**: The top fifty single-word terms in the peach corpus.

Lin *et al*. (2020) conducted a non-experimental descriptive study on health guidelines referred to as patient educational material and fact sheets. They examined educational materials for diabetes patients that were initially written in English by Diabetes Australia and the National Diabetes Services Scheme. They were then translated into multiple languages, including Arabic, for use in Australia. The study aimed to determine if the Arabic-translated materials matched the 8th-grade reading level suggested by health literacy guidelines. By applying readability indices like the Flesch Kincaid Grade Level, Gunning Fog Score, Coleman Liau Index, Simplified Measure of Gobbledygook Index, and the Automated Readability Index (ari), and analysing the average number of syllables per word and words per sentence, Lin *et al*. discovered that the translated materials had a readability level above Grade 10, indicating a higher complexity than the recommended level.

In her study, Sharkas (2019) examined the readability issues of English-to-Arabic translated package inserts, using Jensen's (2013) findings on Danish translations as a foundation. Jensen noted a preference for English over Danish package inserts among the Danish public, suggesting the translations were overly complex and lengthy. Whilst Sharkas did not specifically confirm whether the Arabic-speaking audience found the Arabic translations more challenging than the English originals, she assumed this to be likely and explored the underlying causes. Her analysis of twenty translated inserts indicated that the readability and lay-friendliness issues could be due to the direct use of terms from medical dictionaries, such as translating 'endemic goiter' directly as دراق متوطن. She suggested amplifying or adding descriptive words to clarify terms that could enhance understanding. For example, she recommended expanding 'endemic goiter' to تضخم الغدة الدرقية المتوطن to make it clearer. Sharkas emphasised that translator training should include strategies to maintain simplicity and clarity in translations, suggesting that adding explanations for complex medical terms could improve the accessibility of patient information without sacrificing accuracy.

Malik *et al*. (2019) examined the readability and lay-friendliness of 205 English and Arabic patient educational materials and fact sheets collected from the Medline Plus portal. They used different metrics to assess readability for each language: for English, they used Flesch-Kincaid, and for Arabic, they used the Open-Source Metric for Measuring Arabic Narratives. Their results show that, on average, the texts have fairly good readability in English and Arabic, with Arabic documents sometimes exhibiting better readability than English documents.

The peach corpus stands as a more robust tool for analysing the readability and user-friendliness of patient information leaflets and educational materials, offering advantages over corpora utilised in research by Algabbani *et al*. (2022), Lin *et al*. (2020), Malik *et al*. (2019) and Sharkas (2019). With its collection of 517 files, the peach corpus surpasses the size and scope of Sharkas's (2019) twenty leaflets and Malik *et al*.'s (2019) 205 educational materials, providing a broader base for analysis. Its diversity is notable, including a mix of patient information leaflets and educational materials covering various medications, in contrast to Algabbani *et al*. (2022), who focussed solely on anti-hypertensive medication leaflets. Furthermore, the peach corpus serves multiple research purposes, such as bilingual lexicon extraction, adapting language models for specific sectors, assessing machine translation in healthcare, and aiding translation education. Its manual sentence alignment sets it apart from the corpora in the studies above. Whilst the peach corpus is publicly accessible, enhancing its utility, the availability of the other corpora is not as transparent, though they may be obtained through author correspondence.

Moreno-Sandoval and Campillos-Llanos (2013) compiled a corpus similar to peach. This corpus focusses on patient educational materials in multiple languages (Japanese, Spanish and Arabic). The Arabic section is compiled from Altibbi, a Jordanian medical website resembling Healthline in the US. In addition, it includes articles from the health sections of three Arabic newspapers: *Al-Aswat* (Saudi Arabia), *Youm7* (Egypt) and *El Khabar* (Algeria). This corpus is sizeable, containing approximately 2.5 million Arabic words. The key distinction between this corpus and the peach corpus lies in their alignment. The Moreno-Sandoval and Campillos-Llanos (2013) corpus is comparable, whilst the peach corpus is parallel.

## 3.    Methods

### 3.1    Corpus compilation and preprocessing

Patient information leaflets were provided in html format on the sfda website, with separate pages for English and Arabic (see Figure 1). The health guidelines were available on hit as pdf files with two columns: English and Arabic (see Figure 2). The html and pdf files were collected using an in-house Web crawler. The html files were then parsed using Beautiful Soup (version 4.12.0). At the same time, the pdf files were converted using Sotoor,[5] an Optical Character Reader (ocr) with an Application Programming Interface (api) for batch processing.

---

[5] See: https://api.sotoor.ai/en/home.

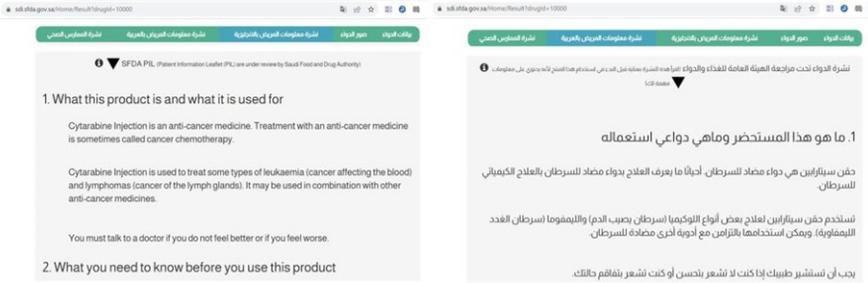

**Figure 1**: A snippet of the html webpages of the patient information leaflets on sfda.

<div dir="rtl">

# العض والخدش بواسطة الحيوانات

</div>

## Animal Bites and Scratches

If you or your child is bitten or scratched by an animal, the wound can get infected. Clean the wound right away and **get medical help as soon as possible**. Even if the animal is your family pet, you should follow these steps:

1. Wash the wound well with soap and water.

2. Put pressure on the area to stop the bleeding.

3. When bleeding stops, put an antibiotic cream, such as Neosporin, on the wound.

4. Cover the bite or scratch with a clean

<div dir="rtl">

إذا تعرضت أنت أو طفلك للعض أو الخدش من قبل أحد الحيوانات، فقد يتعرض الجرح للعدوى. احرصي يعلى تنظيف الجرح على الفور **والحصول على المساعدة الطبية بأسرع ما يمكن**. ينبغي عليك اتباع الخطوات التالية حتى لو كان الحيوان من الحيوانات الأليفة للأسرة:

١. اغسلي الجرح جيدًا بالماء والصابون.

٢. اضغطي على المنطقة لإيقاف النزيف.

٣. عند توقف النزيف، ضعي مرهمًا مضادًا حيويًا، مثل النيوسبورين، على الجرح.

٤. احرصي على تغطية العضة أو الخدش بضمادة نظيفة.

</div>

**Figure 2**: A snippet of the pdf files of health guidelines on hit.

Following the exclusion of duplicates and corrupted files, 260 documents from the sfda and 257 from hit were randomly selected. The initial clean-up process involved the removal of redundant spaces, symbols such as 'tm' and '©', and inconsistent Arabic diacritics. To mitigate the common inaccuracies of ocr technology, thorough manual proofreading was conducted to rectify typographical mistakes and inaccurately converted characters. This proofreading was carried out by senior undergraduate students specialising in translation at the Department of Foreign Languages, University of Sharjah, United Arab Emirates, all of whom are native Arabic speakers.

## 3.2    Corpus alignment

Four Arabic-speaking annotators, all alumni of the translation programme at the Foreign Languages Department, University of Sharjah, United Arab

Emirates, conducted the manual alignment. These annotators were well-versed in different translation disciplines, having taken courses in media, legal, business, literary and medical translation. In addition, each annotator completed a one-semester internship in a translation-centric role and possessed one to two years of experience in professional translation.

The alignment guidelines established for this task were two-fold. Initially, English was designated as the source language and Arabic as the target language – a decision underpinned by the prevalence of English in medical education across the Arab World, except Syria, where medical education is conducted in Arabic (Argeg, 2015; Alshareef *et al.*, 2018; and Sallam, 2021). Subsequently, annotators were briefed on the various alignment types, as detailed in Table 4. These included one-to-one alignment, where a single English sentence corresponds to one Arabic sentence; one-to-many alignment, where a single English sentence translates into multiple Arabic sentences, demarcated by significant punctuation like a full stop; and many-to-one alignment, where multiple English sentences converge into a single, more extended Arabic sentence, typically connected by a co-ordinating or sub-ordinating conjunction or a linking word.

The annotation workflow was methodically organised as follows: Initially, six pairs of annotators (ab, ac, ad, bc, bd and cd) were formed from the four annotators, with no pairings repeated. Each pair was assigned to align eighty-six or eighty-seven texts. To maintain objectivity, annotators worked independently without being provided with the contact information of their counterparts, preventing any potential bias from communication. In the adjudication phase, a systematic approach was employed to address disagreements, where annotators were assigned to review conflicts in another pair's annotations (for instance, A reviewing bc's work and b assessing ac's). After all pairs had completed their annotations and conducted their reviews, I acted as the final adjudicator. In this role, I resolved discrepancies and ensured the comprehensive and accurate alignment of all sentences, ensuring no text was neglected or misaligned.

## 4.    Corpus insights

Table 5 presents statistics from the peach dataset, including counts of word tokens and word types and the Type–Token Ratio (ttr). Table 6 also details the number of sentence pairs in the corpus and their distribution based on the number of words per sentence. The average sentence length in the English part of the corpus ranges from 10.23 to 11.83 words, whilst in the Arabic part, it ranges from 9.52 to 11.41 words.

To delve deeper into the content of the peach corpus, I utilised the keyword extraction function of Sketch Engine (2024). It employs linguistic and statistical methods to compare the corpus under investigation with its

| English | Arabic | Alignment Type |
|---|---|---|
| The allergic reaction might lead to swelling of the face and neck and/or difficulty in breathing or swallowing (Quincke's oedema). | قد يؤدي رد الفعل التحسسي إلى تورم في الوجه والرقبة و/ أو صعوبة في التنفس أو البلع (وذمة كوينكة = وذمة وعائية عصبية). | One-to-one |
| Allergic reactions and increase of connective tissue (fibrosis) in the lung, inflammation of the lining of the lungs or lung scarring (symptoms include coughing, bronchial cramps (bronchospasm), chest discomfort or pain on breathing, breathing difficulties, bloody and/or excessive phlegm); | ردود فعل تحسسية وزيادة النسيج الضام (تليّف) في الرئة، والتهاب في بطانة الرئتين أو تندب الرئة (تشمل الأعراض السعال، تقلص الشعب الهوائية (التشنج القصبي)، وعدم الراحة في الصدر أو ألم مع التنفس، وصعوبات في التنفس، وظهور دم بالبلغم أو زيادة إنتاج البلغم)؛ | One-to-one |
| The repeated administration of Nomal capsules during pregnancy may lead to habituation in the unborn child, and as a result, the child may experience withdrawal symptoms after birth. | قد يؤدي التناول المتكرر لكبسولات نومال أثناء الحمل إلى تعود الطفل الغير مولود عليه. ونتيجة لذلك قد يؤدي هذا إلى ظهور أعراض الانسحاب عند الطفل بعد الولادة. | One-to-many |
| If you feel that your reactions are affected, do not drive a car or another vehicle, do not use electric tools or operate machinery, and do not work without a firm hold! | لا تقم بقيادة السيارة أو المركبات الأخرى، ولا تستخدم الأدوات الكهربائية أو تشغل الآلات إذا شعرت بضعف ردود الفعل لديك. لا تعمل دون أن يكون لديك ما يضمن ثباتك. | One-to-many |
| It is one disease in a group of lung diseases called COPD or chronic obstructive pulmonary disease. The damage often gets worse over time and cannot be cured. | وهو أحد الأمراض في مجموعة أمراض الرئة التي تسمى مرض الانسداد الرئوي المزمن COPD وغالباً ما يزداد المرض سوءاً بمرور الوقت ولا يمكن شفاؤه. | Many-to-one |
| You should not breastfeed while taking Lamisil tablets because your baby would be exposed to terbinafine through your breast milk. This might harm your baby. | يجب أن تمتنعي عن الإرضاع أثناء استعمال لاميزيل أقراص، وذلك لأن طفلك سيتعرّض لتيربينافين من خلال لبن ثديك، وهذا قد يسبب الأذى لطفلك. | Many-to-one |

**Table 4**: Examples of alignment types in the peach corpus.

built-in general-language corpora. The specific settings employed for term extraction were:

*(i)* **Focus on rare words**. This option directs Sketch Engine to highlight rare or unique words within the corpus to identify specialised terminology.

*(ii)* **Only alphanumeric**. This filter ensures that extracted terms are composed only of letters, numbers and hyphens.

|        | English | | | Arabic | | |
|--------|---------|---------|-----|---------|---------|-----|
|        | Tokens  | Types   | TTR | Tokens  | Types   | TTR |
| SFDA   | 456,874 | 24,788  | 0.054 | 440,360 | 42,415 | 0.096 |
| HIT    | 133,643 | 10,966  | 0.082 | 124,347 | 18,690 | 0.15 |
| Total  | 590,517 | 30,554  | 0.19  | 567,707 | 52,462 | 0.11 |

**Table 5**: The peach corpus counts of word tokens, types and type–token ratios.

|        |         | Sentence length | | | | | Total |
|--------|---------|--------|-----------------|-----------------|-------------|---------|-------|
|        |         | 1 word | 2 to 4 words    | 5 to 7 words    | 7+ words    | Average |       |
| SFDA   | English | 1,554  | 6,318           | 6,241           | 24,496      | 11.83   | 38,609 |
|        | Arabic  | 1,239  | 6,910           | 6,668           | 23,801      | 11.41   |       |
| HIT    | English | 417    | 2,155           | 2,294           | 8,196       | 10.23   | 13,062 |
|        | Arabic  | 393    | 2,368           | 2,660           | 7,641       | 9.52    |       |
|        |         |        |                 |                 |             | Total   | 51,671 |

**Table 6**: The peach corpus sentence length distribution.

*(iii)*    **A=a**. This configuration uses a lowercase version of the corpus for case-insensitive results, applicable only to English.

*(iv)*    **English reference corpus**. The reference corpus for English was enTenTen21, with over 52 billion words.

*(v)*    **Arabic reference corpus**. For Arabic, the reference was arTenTen18, containing approximately 4.7 billion words.

Table 7 lists the top fifty single-word terms from the English and Arabic portions of the peach corpus, ranked according to their termhood score. This score evaluates a word's domain specificity by comparing its frequency in the peach corpus against its occurrence in the respective reference corpus. Furthermore, Appendices A and B list some topics derived from the sfda and hit files included in the peach corpus, providing further context and insights into its contents.

## 5.    Corpus uses

As highlighted in Section 1, the peach corpus, being sentence-aligned and parallel, has multi-faceted applications. It holds pedagogical value, particularly for Arab universities that offer courses in medical translation.

| English | *mirzagen, lopresor, mitoxantron, enoxa, nervax, oxetine, pregadex, paxitab, lejam, megamox, olazine, ebewe, latanocom, livazo, nordilet, laprix, urilax, lotevan, mofetab, orotix, omeral, lorvast, neuroplex, linopril, micardisplus, lonzet, lisino, uritab, riapanta, lowrac, prometin, phentolep, loraday, latano, lotense, loradad, metfor, normoten, logynon, mycoheal, mesporin, methergin, lercadip, methotrexat, monozide, optilone, no-uric, prosta-tab, pentolate, oflam* |
|---|---|
| Arabic | بريسيدكس، انوكسا، ايبيفيه، اولمينتك، ايميجران، مايفورتيك، لتانوكوم، نيرڤاكس، نورديلت، اوروتكس، نيزورال، مترونيدازول، لوتيڤْان، بروميتين، يوربلاكس، اتورڤْا، لينوبريل، نوفورابد، مثرجين، موفيتاب، لوجينون، اوميرال، لوراداد، ڤينتولين، يوربتاب، بروجراف، بريتيراكس، أوبتيمول، لورڤْاست، نومال، ميسبورين، بنتاسا، طبيبيك، بلازمكس، مير زاجن، اورينسيا، نوراكتون، نافيلبين، أوبتيبرد، ميجيون، نوڤوسفن، مودوربتيك، لوراداي، يونيڤيت، ميريونال، لورين، الصيدلي، لورينكس |

**Table 7**: Word clouds of the top fifty single-word terms in the peach corpus.

The under-utilisation of parallel corpora in these courses, as Zaki (2021) noted, is often due to their scarcity or limited accessibility. Also, peach provides a resource to corroborate the research findings of Algabbani *et al*. (2022), Lin *et al*. (2020), Malik *et al*. (2019) and Sharkas (2019) regarding the disparities in readability and lay-friendliness between English and Arabic patient information leaflets and educational materials. peach's extensive size and thematic diversity offer a more robust corpus for such comparative analyses.

The corpus also serves as a tool for examining concerns over machine translation in healthcare, as discussed by Das *et al*. (2019), Khoong *et al*. (2019) and Vieira *et al*. (2021), particularly the risks of meaning distortion and the omission or addition of crucial information. Moreover, peach is instrumental in fine-tuning large language models for specific-domain translation and enhancing machine translation systems. An instance of this application is my use of Google AutoML Translation to tailor Google's neural machine translation using the peach corpus. The adaptation increased performance from a bleu score of 31.56, categorised by Google Cloud (2024) as ranging from understandable to good, to 35.7. Despite remaining within the same bleu score bracket, this improvement of 4.14 points is a stride toward enhanced translation quality.

## 6.    Limitations of the corpus

The peach corpus's primary limitation lies in the incomplete details regarding the translation process for the patient information leaflets by the sfda. Whilst Section 1 outlines that the sfda, under the directives of the gcc, mandates the availability of these leaflets in Arabic for public distribution (with no such requirement for hospital-only distributions), it only broadly states that the language should be clear and accurate. Details on the specific translation guidelines, dictionaries utilised or the translators' qualifications remain unspecified. According to the sfda's regulations, pharmaceutical companies are responsible for translating the leaflets and submitting these translations along with the original English versions for sfda's approval (referenced in Section 1). An online service is available for this submission and approval process.[6] Given the vast number of pharmaceutical companies, determining the precise translation methodology for each is challenging. Nonetheless, as sfda is a regulatory authority, the translations it approves within its drug information system can be considered to be reliable reference translations.

## 7.    Conclusion

The peach corpus, introduced in this paper, serves as a novel linguistic resource with diverse applications. As a sentence-aligned parallel corpus comprising patient information leaflets and educational materials, peach holds value across multiple fields, including linguistics, translation studies, translation pedagogy, and machine translation. peach is publicly accessible on Mendeley Data.[7]

## Acknowledgments

T his research was supported by Seed Research Grant No. 2203020129 from the University of Sharjah, United Arab Emirates.

---

[6] See: https://www.sfda.gov.sa/en/eservices/68699.
[7] See: https://data.mendeley.com/datasets/5k6yrrhng7/1.

**Appendix A**: A sample of the topics in the sfda section of the peach corpus.

| File no. | Medication name | File no. | Medication name |
|---|---|---|---|
| 610 | Amantadine sulfate | 824 | Lamisil |
| 284 | Amoxycillin | 847 | Lanzor |
| 11591 | Atrova | 845 | Laprix |
| 11226 | Batlor | 809 | Lasix |
| 374 | Betamethasone | 795 | Latano |
| 557 | Caplet | 794 | Latanoprost |
| 303 | Cefdinir | 783 | Lejam |
| 11617 | Enoxa | 799 | Lendomax |
| 11585 | Foscarnet TBM | 800 | Lercadip |
| 4309 | Imigran | 815 | Levemir |
| 372 | Ivabradine | 811 | Levitra |
| 11598 | Kerasal | 810 | Levonic |
| 830 | Lamictal | 807 | Levophed |
| 829 | Lamifen | 805 | Levox |

**Appendix B**: A sample of the topics in the hit section of the peach corpus.

| File no. | Topic |
|---|---|
| 1 | Bronchitis |
| 2 | Advance Directives |
| 3 | Home Care after Total Joint Replacement |
| 4 | Home Care Instructions after Surgery |
| 5 | Your Hospital Care after Surgery |
| 6 | Angina |
| 7 | Substance Abuse or Dependence |
| 8 | Allergies |
| 9 | Alzheimer's Disease |
| 10 | Barium Enema |
| 11 | Anemia |
| 12 | Nitroglycerin |
| 13 | Heart Cath and Heart Angioplasty |
| 14 | Animal Bites and Scratches |
| 15 | Ankle Exercises |
| 16 | Ankle Sprain |
| 17 | Anthrax: What You Need to Know |
| 18 | Appendectomy |
| 19 | Simple Appendectomy for a Child |
| 20 | Active Range of Motion Exercises: Wrists, Elbows, Forearms, and Shoulders |
| 21 | Assisted Arm Range of Motion Exercises |
| 22 | Wearing a Shoulder Sling |
| 23 | Electrocardiogram (ECG or EKG) |
| 24 | EPS (Electrophysiology Study)s |

**Your short guide to the EUP Journals Blog** http://euppublishingblog.com/

*A forum for discussions relating to Edinburgh University Press Journals*

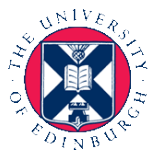

## 1. The primary goal of the EUP Journals Blog

To aid discovery of authors, articles, research, multimedia and reviews published in Journals, and as a consequence contribute to increasing traffic, usage and citations of journal content.

## 2. Audience

Blog posts are written for an educated, popular and academic audience within EUP Journals' publishing fields.

## 3. Content criteria - your ideas for posts

We prioritize posts that will feature highly in search rankings, that are shareable and that will drive readers to your article on the EUP site.

## 4. Word count, style, and formatting

- Flexible length, however typical posts range 70-600 words.
- Related images and media files are encouraged.
- No heavy restrictions to the style or format of the post, but it should best reflect the content and topic discussed.

## 5. Linking policy

- Links to external blogs and websites that are related to the author, subject matter and to EUP publishing fields are encouraged, e.g.to related blog posts

## 6. Submit your post

Submit to ruth.allison@eup.ed.ac.uk

If you'd like to be a regular contributor, then we can set you up as an author so you can create, edit, publish, and delete your *own* posts, as well as upload files and images.

## 7. Republishing/repurposing

Posts may be re-used and re-purposed on other websites and blogs, but a minimum 2 week waiting period is suggested, and an acknowledgement and link to the original post on the EUP blog is requested.

## 8. Items to accompany post

- A short biography (ideally 25 words or less, but up to 40 words)
- A photo/headshot image of the author(s) if possible.
- Any relevant, thematic images or accompanying media (podcasts, video, graphics and photographs), provided copyright and permission to republish has been obtained.
- Files should be high resolution and a maximum of 1GB
- Permitted file types: *jpg, jpeg, png, gif, pdf, doc, ppt, odt, pptx, docx, pps, ppsx, xls, xlsx, key, mp3, m4a, wav, ogg, zip, ogv, mp4, m4v, mov, wmv, avi, mpg, 3gp, 3g2.*